\documentclass{article}

\usepackage[preprint]{nips_2018}

\usepackage[utf8]{inputenc} 
\usepackage[T1]{fontenc}    
\usepackage{hyperref}       
\usepackage{url}            
\usepackage{booktabs}       
\usepackage{amsfonts}       
\usepackage{nicefrac}       
\usepackage{microtype}      
\usepackage{xspace}
\usepackage{amsthm}
\usepackage{amsmath}
\usepackage[ruled,vlined,linesnumbered,algo2e]{algorithm2e}
\usepackage{natbib}
\usepackage{tkz-graph}
\usepackage{subcaption}
\usepackage{siunitx}
\usepackage{booktabs}
\usepackage{wrapfig}

\title{Approximate Knowledge Compilation by\\  Online Collapsed Importance Sampling}

\author{
  Tal Friedman \\
  Computer Science Department\\
  University of California\\
  Los Angeles, CA 90095 \\ 
  \texttt{tal@cs.ucla.edu} \\
  \And
  Guy Van den Broeck \\
  Computer Science Department\\
  University of California\\
  Los Angeles, CA 90095 \\ 
  \texttt{guyvdb@cs.ucla.edu} \\
}
\usepackage{tikz}
\usetikzlibrary{arrows,shapes,shapes.geometric,backgrounds,arrows.meta}
\usetikzlibrary{trees}
\usetikzlibrary{calc}
\usetikzlibrary{decorations.markings}
\usetikzlibrary{fit}

\tikzstyle{graph}=[
  >=stealth,font=\normalsize,auto,scale=1,every node/.style={scale=1}
]
\tikzstyle{diagram}=[
  >=stealth,font=\small,auto,scale=0.9,every node/.style={scale=0.9}
]
\tikzstyle{node}=[
  draw,circle,inner sep=3pt,fill=white
]
\tikzstyle{box}=[
  draw,rectangle,inner sep=3pt,fill=white
]
\tikzstyle{nnf}=[
  >=stealth,font=\footnotesize,auto,scale=0.7,every node/.style={scale=0.7}
]
\tikzstyle{bn}=[
  >=stealth,font=\small,auto,scale=.9,every node/.style={scale=.9}
]
\tikzstyle{bn2}=[
  >=stealth,font=\small,auto,scale=1,every node/.style={scale=1}
]
\tikzstyle{extnode}=[
  draw,circle,inner sep=2pt,fill=white
]
\tikzstyle{decnode}=[
  draw,ellipse,inner sep=2pt,fill=white
]
\tikzstyle{leafnode}=[
  draw,rectangle,fill=gray!20,inner sep=3.5pt
]
\tikzstyle{obnode}=[
  draw,ellipse,inner sep=2pt,fill=black!25!white
]
\tikzstyle{dashnode}=[
  draw,dashed,ellipse,inner sep=2pt,fill=white
]
\tikzstyle{dashobnode}=[
  draw,dashed,ellipse,inner sep=2pt,fill=black!25!white
]

\tikzstyle{fgedge}=[
    line width=0.3
]
\tikzstyle{fg}=[
  >=latex, thick, auto
]
\tikzstyle{fgnode}=[
  draw,ellipse,minimum size=7mm,inner sep=1pt,font=\small
]
\tikzstyle{fgfactor}=[
  draw,minimum size=1mm,inner sep=2pt,font=\small, fill=gray!50
]
\tikzstyle{factortable}=[
  font=\footnotesize
]

\tikzstyle{nnf}=[
  >=stealth,font=\small,auto,scale=0.9,every node/.style={scale=0.9}
]
\tikzstyle{extnode}=[
  draw,circle,inner sep=1pt,fill=white, line width=0.25
]
\tikzstyle{leafnode}=[
  draw,fill=gray!20,inner sep=2pt, line width=0.25
]
\tikzstyle{constnode}=[
  draw,fill=white,inner sep=2pt, line width=0.25
]
\tikzstyle{label}=[
  fill=white,inner sep=2.5pt
]

\tikzstyle{acarrow}=[
    decoration={markings,mark=at position 1 with {\arrow[scale=0.6]{>}}},
    postaction={decorate},
    shorten >=0.4pt,
    >=latex, line width=0.25
]

\newtheorem{theorem}{Theorem}
\newtheorem{definition}{Definition}
\newtheorem{corollary}{Corollary}

\newcommand{\rvar}[1]{\ensuremath{\mathit{#1}}\xspace}
\newcommand{\Xr}{\rvar{X}}

\newcommand{\Sr}{\rvar{S}}
\newcommand{\Fr}{\rvar{F}}

\newcommand{\Vr}{\rvar{V}}

\newcommand{\rvars}[1]{\ensuremath{\mathbf{#1}}\xspace}
\newcommand{\Xs}{\rvars{X}}
\newcommand{\Ys}{\rvars{Y}}
\newcommand{\Ss}{\rvars{S}}
\newcommand{\Vs}{\rvars{V}}
\newcommand{\Fs}{\rvars{F}}

\newcommand{\Xp}{\rvars{X_p}}
\newcommand{\Xd}{\rvars{X_d}}

\newcommand{\Xpm}{\rvars{X^m_p}}
\newcommand{\Xdm}{\rvars{X^m_d}}

\newcommand{\xp}{\jstate{x_p}}
\newcommand{\xd}{\jstate{x_d}}

\newcommand{\xpm}{\jstate{x^m_p}}

\newcommand{\Ex}{\mathbb{E}}

\newcommand{\jstate}[1]{\ensuremath{\mathbf{#1}}\xspace}
\newcommand{\xs}{\jstate{x}}
\newcommand{\ys}{\jstate{y}}

\DeclareMathOperator*{\argmin}{argmin}

\DeclareMathOperator*{\dist}{distance}

\begin{document}

\maketitle
\begin{abstract}
We introduce collapsed compilation, a novel approximate inference algorithm for discrete probabilistic graphical models. 
It is a collapsed sampling algorithm that incrementally selects which variable
to sample next based on the partial sample obtained so far. This online
collapsing, together with knowledge compilation inference on the remaining
variables, naturally exploits local structure and context-specific independence
in the distribution. These properties are naturally exploited in exact inference, but are difficult to harness for approximate inference.
Moreover, by having a partially compiled circuit available
during sampling, collapsed compilation has access to a highly effective proposal
distribution for importance sampling. Our experimental evaluation shows that
collapsed compilation performs well on standard benchmarks. In particular, when the
amount of exact inference is equally limited, collapsed
compilation is competitive with the state of the art, and outperforms it on several benchmarks.
\end{abstract}

\section{Introduction}

Modern probabilistic inference algorithms for discrete graphical models are designed to exploit key properties of the distribution. In addition to classical conditional independence, they exploit local structure in the individual factors, determinism coming from logical constraints~\citep{darwiche2009modeling}, and the context-specific independencies that arise in such distributions~\citep{boutilier1996context}.
The \emph{knowledge compilation} approach in particular forms the basis for state-of-the-art probabilistic inference algorithms in a wide range of models, including Bayesian networks~\citep{Chavira2008OnPI}, factor graphs~\citep{Choi2013CompilingPG}, relational models~\citep{chavira2006compiling}, probabilistic programs~\citep{fierens2015inference}, probabilistic databases~\citep{VdBFTDB17}, and dynamic Bayesian networks~\citep{VlasselaerAIJ16}.
Based on logical reasoning techniques, knowledge compilation algorithms construct an \emph{arithmetic circuit} representation of the distribution on which inference is guaranteed to be efficient~\citep{Darwichediff2003}.
The inference algorithms listed above have one common limitation: they perform exact inference by compiling a worst-case exponentially-sized arithmetic circuit representation.
Our goal in this paper is to upgrade these techniques to allow for \emph{approximate probabilistic inference}, while still naturally exploiting the structure in the distribution. We aim to open up a new direction towards scaling up knowledge compilation to larger distributions.

When knowledge compilation produces circuits that are too large, a natural solution is to sample some random variables, and to do exact compilation on the smaller distribution over the remaining variables. This collapsed sampling approach suffers from two problems. First, collapsed sampling assumes that one can determine a priori which variables need to be sampled to make the distribution amenable to exact inference. When dealing with large amounts of context-specific independence, it is difficult to find such a set, because the independencies are a function of the particular values that variables get instantiated to. 
Second, collapsed sampling assumes that one has access to a proposal distribution for the sampled variables, and the success of inference largely depends on the quality of this proposal. In practice, the user often needs to specify the proposal distribution, and it is difficult to automatically construct one which is general-purpose.

As our first contribution, Section~\ref{s:online} introduces \emph{online collapsed importance sampling}, where the sampler chooses which variable to sample next based on the values sampled for previous variables. This algorithm is a solution to the first problem identified above. We show that the sampler corresponds to a classical collapsed importance sampler on an augmented graphical model and prove conditions for it to be asymptotically unbiased.

Section~\ref{sec:colcomp} describes our second contribution: a \emph{collapsed compilation} algorithm that maintains a partially compiled arithmetic circuit during online collapsed sampling. This circuit serves as a highly-effective proposal distribution at each step of the algorithm. By setting a limit on the circuit size as we compile more factors into the model, we are able to sample exactly as many variables as needed to fit the arithmetic circuit in memory. Moreover, through online collapsing, the set of collapsed variables changes with every sample, exploiting different independencies in each case.

Finally, we experimentally validate the performance of collapsed compilation on
standard benchmarks. We begin by empirically examining properties of collapsed compilation, to show the value of the proposal distribution and
pick apart where performance improvements are coming from. Then, in a setting where the amount of exact inference is fixed, we find that collapsed compilation is competitive with
state-of-the-art approximate inference algorithms, outperforming them on several benchmarks. 

\section{Online Collapsed Importance Sampling} \label{s:online}


We begin with a brief review of collapsed
importance sampling, before motivating the need for dynamically
selecting which variables to sample. We then demonstrate that we can select
variables in an online fashion while
maintaining the desired unbiasedness property, using an algorithm we call online
collapsed importance sampling.

We denote random variables with uppercase letters ($\Xr$), and their instantiation
with lowercase letters ($x$). Bold letters denote  sets
($\Xs$) and their instantiations~($\xs$).
We refer to \citet{koller2009probabilistic} for notation
and formulae related to (collapsed) importance sampling. 

\subsection{Collapsed Importance Sampling}
The basic principle behind collapsed sampling is that we can reduce the variance
of an estimator by making part of the inference exact. That is, suppose we
partition our variables into two sets: $\Xp$, and $\Xd$. 
In collapsed importance sampling, the distribution of variables in $\Xp$
will be estimated via importance sampling, while those in $\Xd$ will be
estimated by computing exactly $P(\Xd | \xp)$ for each sample $\xp$. In
particular, suppose we have some function $f(\xs)$ where $\xs$ is a complete
instantiation of $\Xp \cup \Xd$, and a proposal distribution $Q$ over $\Xp$. Then we estimate the expectation of $f$ by
\begin{equation}
  \label{eq:colimsamp}
  \hat{\mathbb{E}}(f) = \frac{\sum_{m=1}^M w[m] (\mathbb{E}_{P(\Xd | \xp[m])}[f(\xp[m], \Xd)])}{\sum_{m=1}^Mw[m]},
\end{equation}
where we have drawn samples $\{\xp[m]\}_{m=1}^M$ from a proposal distribution $Q$,
and for each sample analytically computed the importance weights
$w[m]=\frac{P(\xp[m])}{Q(\xp[m])}$, as well as $\mathbb{E}_{P(\Xd |
  \xp[m])}[f(\xp[m], \Xd)]$. Due to the properties of importance samplers, the
estimator given by (\ref{eq:colimsamp}) is asymptotically unbiased. Moreover, if we can
compute $P(\xp[m])$ exactly rather than the unnormalized $\hat{P}(\xp[m])$, then
the estimator is unbiased \citep{tokdar2010importance}.

\subsection{Motivation} \label{s:motivation}
A critical decision that needs to be made when doing collapsed sampling is selecting a
partition -- that is, which variables go in $\Xp$ and which go in $\Xd$. The
choice of partition can have a large effect on the quality of the resulting
estimator, and the process of choosing such a partition requires expert
knowledge. Furthermore, selecting a partition a priori
that works well is not always possible, as we will show in the following
example. All of this raises the question whether it is possible to choose the
partition on the fly for each sample, which we will discuss in Section~\ref{sec:onlalg}.

\begin{figure}
\centering
\begin{subfigure}[b]{.24\textwidth}
  \centering
  \begingroup
    \renewcommand*{\arraystretch}{1.1}
      \begin{tabular}{|l|l|}
        \hline
        $F_{12}$ & 1 \\
        \hline
        $F_{13}$ & 0 \\
        \hline
        $F_{23}$ & 1 \\
        \hline
      \end{tabular} 
    \endgroup
  \caption{Sampled Friendships}
  \label{subfig:mota}
\end{subfigure}%
\begin{subfigure}[b]{.24\textwidth}
  \centering
  \scalebox{0.78}{
  \begin{tikzpicture}
      \Vertex[Math,L=V_1,x=2, y=2]{v1}
      \Vertex[Math,L=V_2,x=1, y=1]{v2}
      \Vertex[Math,L=V_3,x=3, y=1]{v3}
      \Edge(v1)(v2)
      \Edge(v2)(v3)
  \end{tikzpicture}  } 
  \caption{Induced Dependencies}
  \label{subfig:motb}
\end{subfigure}%
  \begin{subfigure}[b]{.24\textwidth}
       \centering
    \begin{tikzpicture}[scale=0.6, every node/.style={transform shape}]
      \Vertex[Math,L=V_1,x=2, y=5]{v1}
      \Vertex[Math,L=V_2,x=3, y=5]{v2}
      \Vertex[Math,L=V_3,x=1, y=4]{v3}
      \Vertex[Math,L=V_4,x=4, y=4]{v4}
      \Vertex[Math,L=V_5,x=2, y=3]{v5}
      \Vertex[Math,L=V_6,x=3, y=3]{v6}
      \Vertex[Math,L=V_7,x=2, y=2]{v7}
      \Vertex[Math,L=V_8,x=1, y=2]{v8}
      \Vertex[Math,L=V_9,x=3, y=2]{v9}
      \Vertex[Math,L=V_{10},x=4, y=2]{v10}
      \Edge(v1)(v2)
      \Edge(v1)(v3)
      \Edge(v1)(v4)
      \Edge(v1)(v5)
      \Edge(v1)(v6)
      \Edge(v2)(v3)
      \Edge(v2)(v4)
      \Edge(v2)(v5)
      \Edge(v2)(v6)
      \Edge(v3)(v5)
      \Edge(v4)(v5)
      \Edge(v4)(v6)
      \Edge(v5)(v6)
      \Edge(v7)(v8)
      \Edge(v7)(v9)
      \Edge(v9)(v10)
    \end{tikzpicture}
    \caption{Induced Network $G_1$}
    \label{subfig:motc}
    \end{subfigure}%
 \begin{subfigure}[b]{.24\textwidth}
  \centering
    \begin{tikzpicture}[scale=0.6, every node/.style={transform shape}]
      \Vertex[Math,L=V_1,x=3.5, y=4]{v1}
      \Vertex[Math,L=V_2,x=2.5, y=4]{v2}
      \Vertex[Math,L=V_3,x=4.5, y=4]{v3}
      \Vertex[Math,L=V_4,x=1.5, y=4.0]{v4}
      \Vertex[Math,L=V_5,x=1.5, y=3.0]{v5}
      \Vertex[Math,L=V_6,x=3, y=3]{v6}
      \Vertex[Math,L=V_7,x=2, y=2]{v7}
      \Vertex[Math,L=V_8,x=4, y=2]{v8}
      \Vertex[Math,L=V_9,x=2.5, y=1]{v9}
      \Vertex[Math,L=V_{10},x=3.5, y=1]{v10}
      \Edge(v1)(v2)
      \Edge(v1)(v3)
      \Edge(v2)(v4)
      \Edge(v2)(v5)
      \Edge(v6)(v7)
      \Edge(v6)(v8)
      \Edge(v6)(v9)
      \Edge(v6)(v10)
      \Edge(v7)(v8)
      \Edge(v7)(v10)
      \Edge(v8)(v9)
      \Edge(v8)(v10)
      \Edge(v9)(v10)
    \end{tikzpicture}
    \caption{Induced Network $G_2$}
    \label{subfig:motd}
\end{subfigure}
\caption{Different samples for $\Fs$ can have a large effect on the resulting dependencies between $\Vs$.}
  \label{fig:motivation}
\end{figure}
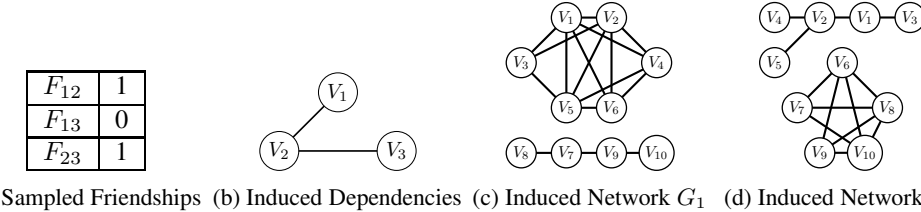

Suppose we have a group of $n$ people, denoted ${1,...,n}$. For every pair of
people $(i,j), i < j$, there is a binary variable $\Fr_{ij}$ indicating whether $i$
and $j$ are friends. Additionally, we have
features $V_i$ for each person $i$, and $F_{ij}=1$ (that is $i,j$ are friends)
implies that $\Vr_i$ and $\Vr_j$ are correlated.
Suppose we are performing collapsed sampling on the joint distribution over $\Fs, \Vs$, and that we have already decided to place all friendship indicators
$\Fr_{ij}$ in $\Xp$ to be sampled. Next, we need to decide which variables in $\Vs$ to include in $\Xp$ for the remaining inference problem over $\Xd$ to become tractable.
Observe that given a sampled $\Fs$, due to the independence
properties of $\Vs$ relying on $\Fs$, a graphical model $G$ is induced
over $\Vs$ (see Figures~\ref{subfig:mota},\ref{subfig:motb}). Moreover, this graphical model can vary greatly between different
samples of $\Fs$. For example, $G_1$ in Figure~\ref{subfig:motc} densely connects $\{\Vr_1,...,\Vr_6\}$ making it difficult to perform exact inference. Thus, we
will need to sample some variables from this set. However, exact
inference over $\{\Vr_7,...,\Vr_{10}\}$ is easy. Conversely, $G_2$ in Figure~\ref{subfig:motd} depicts the  opposite scenario: $\{\Vr_1,...,\Vr_5\}$ forms a tree, which is easy for inference, whereas $\{\Vr_6,...,\Vr_{10}\}$ is now intractable. 
It is clearly impossible to choose a small
subset of $\Vs$ to sample that fits all cases, thus demonstrating a need for
an online variable selection during collapsed sampling.

\subsection{Algorithm}
\label{sec:onlalg}
We now introduce our online collapsed importance sampling algorithm. It decides at sampling time which variables to sample and which to do exact inference on.

To gain an intuition, suppose we are in the standard collapsed importance sampling setting.
Rather than sampling an instantiation $\xp$ jointly from $Q$, we can instead
first sample $x_{p_1} \sim Q(\Xr_{p_1})$, then $x_{p_2} \sim Q(\Xr_{p_2}|x_{p_1})$, and so on
using the chain rule of probability. In online collapsed importance sampling,
rather than deciding $\Xr_{p_1},\Xr_{p_2},...$ a priori, we select which variable we
will use as $\Xr_{p_2}$ based on the previous sampled value $x_{p_1}$.

\begin{definition}
  Let $\ys$ be an instantiation of $\Ys \subset
  \Xs$. A \emph{variable selection policy} $\pi$ takes $\ys$ and either stops sampling or returns a distribution over $\Xs \setminus \Ys$ that
   decides the next variable to sample.
  \end{definition}
\begin{wrapfigure}[16]{r}{.42\textwidth}
\vspace{-0.5cm}
\begin{flushright}
\begin{minipage}{.970\linewidth}
\begin{algorithm2e}[H]
  \caption{Online Collapsed IS }
  \label{alg:oncolims}
  \SetKwInOut{Input}{Input}
  \DontPrintSemicolon
  \Input{$\Xs$: The set of all variables, \\
    $Q_{\Xr_i}$: Proposal
    distributions, \\ $\pi$: Variable selection policy}
  \KwResult{A sample $\left(\Xdm,\xpm,w[m]\right)$}
  $\xp \leftarrow \{\}$ ;~~ $\Xd \leftarrow \Xs$ \;
  \While{$\pi$ does not stop}{
    $\Xr_i \sim \pi$ ($\xp$) \;
    $x_i \sim Q_{\Xr_i}(\Xr_i | \xp)$ \;
    $\xp \leftarrow \xp \cup \{x_i\}$ \;
    $\Xd \leftarrow \Xd \setminus \{\Xr_i\}$
  }
  \Return{$\left(\Xd, \xp, \frac{\hat{P}(\xp)}{Q(\xp)}\right)$}
  \end{algorithm2e}
  \end{minipage}
  \end{flushright}
\end{wrapfigure}

For example, a naive policy could be to select a remaining variable uniformly at random. Once the policy $\pi$ stops sampling, we are left with an instantiation $\xp$ and a set
$\Xd$, where both  are specific to the particular~sample.

Algorithm~\ref{alg:oncolims} shows more precisely how online collapsed
importance sampling generates a
single sample, given a full set of variables $\Xs$, a variable selection policy
$\pi$, and proposal distributions
$Q_{\Xr_i}$ for any choice of $\Xr_i$ and $\xp$. 
Note that $\xp$ is a set of variables together with their instantiations, while $\Xd$ is just a set of variables.
Notationally, for sample $m$ we write these as
$\xpm$ and~$\Xdm$.
The joint proposal $Q(\xp)$ is left abstract for now (see Section~\ref{s:proposalcompute} for a concrete instance). In general, it is induced by $\pi$ and the individual local proposals $Q_{\Xr_i}$.

\begin{definition}
Given $m$ samples $\left\{\Xdm,\xpm,w[m]\right\}$ produced by online collapsed importance
sampling, the \emph{online
collapsed importance sampling estimator} of $f$ is
\begin{equation}
  \label{eq:onlcolimsamp}
  \hat{\mathbb{E}}(f) = \frac{\sum_{m=1}^M w[m] (\mathbb{E}_{P(\Xdm | \xpm)}[f(\xpm, \Xdm)])}{\sum_{m=1}^Mw[m]}.
\end{equation}

\end{definition}
Note that the only difference compared to Equation~\ref{eq:colimsamp} is
that sets $\Xpm$ and $\Xdm$ vary with each sample.

\subsection{Analysis}

Our algorithm for online collapsed importance sampling raises two questions: does Equation~\ref{eq:onlcolimsamp} yield unbiased estimates, and how does one compute the proposal $Q(\xp)$? We study both questions next.

\subsubsection{Unbiasedness of Estimator}
\begin{figure}[tbh]
  \centering
  \begin{subfigure}{.4\textwidth}
    \centering
    \scalebox{0.8}{
      \begin{tikzpicture}
          \Vertex[Math,L=X_1,x=3, y=3]{x1}
          \tikzset{VertexStyle/.append style={rectangle,fill=gray!50}}
          \Vertex[Math,L=f,x=2, y=3]{f1}
          \tikzset{VertexStyle/.append style={circle,minimum size=40pt,fill=white}}
          \Vertex[Math,L=\mathbf{X_{2:n}},x=0.5,y=3]{xr}
          \Edge(f1)(x1)
          \Edge(f1)(xr)
        \end{tikzpicture}
    }
  \caption{Original factor graph $F$}
  \label{fig:origfg}
\end{subfigure}%
~
\begin{subfigure}{.5\textwidth}
  \centering
  \scalebox{0.8}{
  \begin{tikzpicture}
    \Vertex[Math,L=X_1,x=3, y=3]{x1}
    \Vertex[Math,L=X_1^*,x=5,y=2.55]{xa1}
    \Vertex[Math,L=S_1,x=5,y=3.45]{s1}
      \tikzset{VertexStyle/.append style={rectangle,fill=gray!50}}
      \Vertex[Math,L=f,x=2, y=3]{f1}
      \Vertex[Math,L=f_{au},x=4,y=3]{fau}
      \tikzset{VertexStyle/.append style={circle,minimum size=40pt,fill=white}}
      \Vertex[Math,L=\mathbf{X_{2:n}},x=0.5,y=3]{xr}
      \Edge(f1)(x1)
      \Edge(f1)(xr)
      \Edge(x1)(fau)
      \Edge(fau)(s1)
      \Edge(fau)(xa1)
      
    \end{tikzpicture}
    }
    \caption{Augmented factor graph $F_A$}
  \label{fig:augfg}
\end{subfigure}%
\caption{Online collapsed sampling corresponds to collapsed sampling on an augmented graph \label{fig:augmentingfg}}
\end{figure}
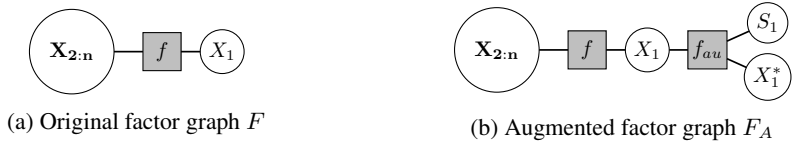

 If we let $\pi$ be a policy
that always returns the same variables in the same order, then we recover
classical offline collapsed importance sampling - and thus retain all of the properties of
offline collapsed importance sampling. With this in mind, we present the augmented factor graph construction given in Figure~\ref{fig:augmentingfg}, which will allow us to make a similar statement for any arbitrary policy $\pi$. 

Intuitively, $\Xr^*_i$ is a copy variable of $\Xr_i$, where we are
sampling $\Xr^*_i$ and computing $\Xr_i$ exactly. To make this possible without
actually inferring the entire distribution exactly, we have $f_{au}$ cause each $\Sr_i$ to act as an
indicator for whether $\Xr^*_i$ and $\Xr_i$ are constrained to be equal. $\Sr_i$ can
also be thought of as indicating whether or not we are sampling $\Xr_i$ in our
original factor graph when doing online collapsed importance sampling. We are now ready to prove the following results.
\begin{theorem}
  \label{thm:equivaug}
For any factor graph $F$ and its augmented graph $F_A$, we have $\forall $\xs$, \Ex_F(\xs) = \Ex_{F_A}(\xs)$.  
\end{theorem}

\begin{theorem}
  \label{thm:onlest}
  Let $F$ be a factor graph and let $F_A$ be its augmented factor graph. Consider the collapsed importance sampling estimator
  (Eq.~\ref{eq:colimsamp}) with $\Xp=\Xs^* \cup \Ss$ and $\Xd=\Xs$. This
  estimator is equivalent to using the online collapsed importance sampling
  estimator (Eq.~\ref{eq:onlcolimsamp}) on~$F$.
\end{theorem}

\begin{corollary}
  The estimator given by Equation \ref{eq:onlcolimsamp} is asymptotically unbiased.
\end{corollary}

Proofs and the details of this construction can be found in Appendix~\ref{a:proofs}.

\subsubsection{Computing the Proposal Distribution} \label{s:proposalcompute}
Our next question is how to compute the global $Q(\xp)$, given that we have each local $Q_{\Xr_{p_i}}(\Xr_{p_i}|\xs_{p_1:i-1})$. Notice that since $Q_{\Xr_{p_i}}$ is not a conditional from $Q$, but rather a generic distribution, this computation is not easy in general. In particular, considering $|\Xp|=n$ and our previous example of a uniformly random policy, then for any given instantiation $\xp$, there
are $n!$ different ways $\xp$ could be sampled from $Q$ - one for each ordering that arrives at $\xp$.
In this case, computing $Q(\xp)$ requires summing over
exponentially many terms, which is undesirable. Instead, we restrict the
variable selection policies we use to the following class, which makes computing $Q(\xp)$ easy.

\begin{definition}
  \label{def:det}
  A \emph{deterministic} variable selection policy $\pi(\xp)$ is a function with a range of $\Xs \setminus \Xp$.
\end{definition}

\begin{theorem}
  \label{thm:prop}
For any sample $\xp$ and deterministic selection policy $\pi(\xp)$, there is exactly one order in which the
variables $\Xp$ could have been sampled. Therefore, $Q(\xp)=\prod_{i=1}^{|\Xp|}Q_{X_{p_i}}(x_{p_i}|\xs_{p_{1:i-1}})$.
\end{theorem}

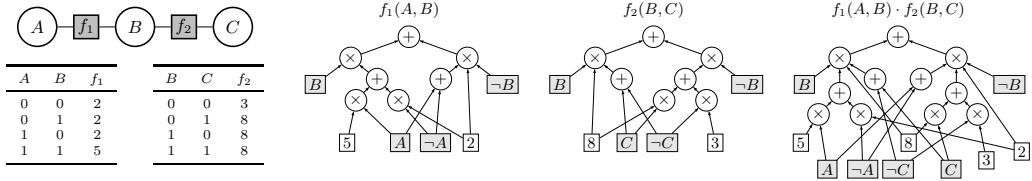
\begin{figure*}
\centering
\scalebox{0.74}{
  \begin{tikzpicture}[fg]
	
    \def\lone{0pt}
    \def\ltwo{-45pt}
    \def\spacing{25pt}

    \node (A) at (0pt,\lone) [fgnode] {$A$};
    \node (f1) at (1*\spacing,\lone) [fgfactor] {$f_1$};
    \node (B) at (2*\spacing,\lone) [fgnode] {$B$};
    \node (f2) at (3*\spacing,\lone) [fgfactor] {$f_2$};
    \node (C) at (4*\spacing,\lone) [fgnode] {$C$};
    
    \begin{scope}[on background layer]
      \draw [fgedge] (A) -- (f1);
      \draw [fgedge] (B) -- (f1);
      \draw [fgedge] (B) -- (f2);
      \draw [fgedge] (C) -- (f2);
    \end{scope}
    
    \node at (0.5*\spacing,\ltwo) [factortable] {
      \scriptsize
      \begin{tabular}{ccc}
	\toprule
	$A$ & $B$ & $f_1$\\\midrule
	$0$ & $0$ & $2$\\
	$0$ & $1$ & $2$\\
	$1$ & $0$ & $2$\\
	$1$ & $1$ & $5$\\
	\bottomrule
      \end{tabular}
    };
    \node at (3.5*\spacing,\ltwo) [factortable] {
      \scriptsize
      \begin{tabular}{ccc}
	\toprule
	$B$ & $C$ & $f_2$\\\midrule
	$0$ & $0$ & $3$\\
	$0$ & $1$ & $8$\\
	$1$ & $0$ & $8$\\
	$1$ & $1$ & $8$\\
	\bottomrule
      \end{tabular}
    };
    
    \begin{scope}[shift={(190pt,-5bp)},nnf]
    
      \node (db) at (0bp,0bp) [extnode] {$+$};
      
      \node (name) at ($(db) + (0bp,+15bp)$) [label] {$f_1(A,B)$};
      
      \node (db1) at ($(db) + (-33bp,-12bp)$) [extnode] {$\times$};
      \node (db0) at ($(db) + (+33bp,-12bp)$) [extnode] {$\times$};
      
      \node (da) at ($(db1) + (+15bp,-12bp)$) [extnode] {$+$};
      \node (da1) at ($(da) + (-12bp,-12bp)$) [extnode] {$\times$};
      \node (da0) at ($(da) + (+12bp,-12bp)$) [extnode] {$\times$};
      
      \node (sa) at ($(db0) + (-15bp,-12bp)$) [extnode] {$+$};
            
      \node (p5) at (-35bp,-60bp) [constnode] {$5$};
      \node (p2) at (35bp,-60bp) [constnode] {$2$};

      \node (b1) at ($(db1) + (-20bp,-15bp)$) [leafnode] {$B$};
      \node (b0) at ($(db0) + (+20bp,-15bp)$) [leafnode] {$\neg B$};
      
      \node (a1) at (-05bp,-60bp)[leafnode] {$A$};
      \node (a0) at (15bp,-60bp) [leafnode] {$\neg A$};
      
      \begin{scope}[on background layer]
        \draw [acarrow] (db0) -- (db);
        \draw [acarrow] (db1) -- (db);
        
        \draw [acarrow] (b0) -- (db0);
        \draw [acarrow] (b1) -- (db1);
        \draw [acarrow] (p2) -- (db0);
        
        \draw [acarrow] (da) -- (db1);
        \draw [acarrow] (da0) -- (da);
        \draw [acarrow] (da1) -- (da);
        
        \draw [acarrow] (a0) -- (da0);
        \draw [acarrow] (a1) -- (da1);
        \draw [acarrow] (p2) -- (da0);
        \draw [acarrow] (p5) -- (da1);
        
        \draw [acarrow] (sa) -- (db0);
        \draw [acarrow] (a0) -- (sa);
        \draw [acarrow] (a1) -- (sa);
        
      \end{scope}
    \end{scope}
    

    \begin{scope}[shift={(315pt,-5bp)},nnf]
    
      \node (db) at (0bp,0bp) [extnode] {$+$};
      
      \node (name) at ($(db) + (0bp,+15bp)$) [label] {$f_2(B,C)$};
      
      \node (db1) at ($(db) + (-33bp,-12bp)$) [extnode] {$\times$};
      \node (db0) at ($(db) + (+33bp,-12bp)$) [extnode] {$\times$};
      
      \node (sc) at ($(db1) + (+15bp,-12bp)$) [extnode] {$+$};
            
      \node (dc) at ($(db0) + (-15bp,-12bp)$) [extnode] {$+$};
      \node (dc1) at ($(dc) + (-12bp,-12bp)$) [extnode] {$\times$};
      \node (dc0) at ($(dc) + (+12bp,-12bp)$) [extnode] {$\times$};
      
      \node (p8) at (-35bp,-60bp) [constnode] {$8$};
      \node (p3) at (35bp,-60bp) [constnode] {$3$};

      \node (b1) at ($(db1) + (-20bp,-15bp)$) [leafnode] {$B$};
      \node (b0) at ($(db0) + (+20bp,-15bp)$) [leafnode] {$\neg B$};
      
      \node (c1) at (-15bp,-60bp)[leafnode] {$C$};
      \node (c0) at (05bp,-60bp) [leafnode] {$\neg C$};
      
      \begin{scope}[on background layer]
        \draw [acarrow] (db0) -- (db);
        \draw [acarrow] (db1) -- (db);
        
        \draw [acarrow] (b0) -- (db0);
        \draw [acarrow] (b1) -- (db1);
        \draw [acarrow] (p8) -- (db1);
        
        \draw [acarrow] (dc) -- (db0);
        \draw [acarrow] (dc0) -- (dc);
        \draw [acarrow] (dc1) -- (dc);
        
        \draw [acarrow] (c0) -- (dc0);
        \draw [acarrow] (c1) -- (dc1);
        \draw [acarrow] (p3) -- (dc0);
        \draw [acarrow] (p8) -- (dc1);
        
        \draw [acarrow] (sc) -- (db1);
        \draw [acarrow] (c0) -- (sc);
        \draw [acarrow] (c1) -- (sc);
        
      \end{scope}
    \end{scope}
    \begin{scope}[shift={(440pt,-5bp)},nnf]
    
      \node (db) at (0bp,0bp) [extnode] {$+$};
      
      \node (name) at ($(db) + (0bp,+15bp)$) [label] {$f_1(A,B) \cdot f_2(B,C)$};
      
      \node (db1) at ($(db) + (-33bp,-12bp)$) [extnode] {$\times$};
      \node (db0) at ($(db) + (+33bp,-12bp)$) [extnode] {$\times$};
      
      \node (sa) at ($(db0) + (-20bp,-12bp)$) [extnode] {$+$};
      \node (sc) at ($(db1) + (+20bp,-12bp)$) [extnode] {$+$};
            
      \node (da) at ($(db1) + (0bp,-20bp)$) [extnode] {$+$};
      \node (da1) at ($(da) + (-12bp,-12bp)$) [extnode] {$\times$};
      \node (da0) at ($(da) + (+12bp,-12bp)$) [extnode] {$\times$};
      
      \node (dc) at ($(db0) + (0bp,-20bp)$) [extnode] {$+$};
      \node (dc1) at ($(dc) + (-12bp,-12bp)$) [extnode] {$\times$};
      \node (dc0) at ($(dc) + (+12bp,-12bp)$) [extnode] {$\times$};
      
      \node (p5) at (-55bp,-60bp) [constnode] {$5$};
      \node (p2) at (70bp,-65bp) [constnode] {$2$};
      \node (p8) at (06bp,-60bp) [constnode] {$8$};
      \node (p3) at (50bp,-70bp) [constnode] {$3$};

      \node (b1) at ($(db1) + (-20bp,-15bp)$) [leafnode] {$B$};
      \node (b0) at ($(db0) + (+30bp,-15bp)$) [leafnode] {$\neg B$};
      
      \node (a1) at (-40bp,-75bp)[leafnode] {$A$};
      \node (a0) at (-20bp,-75bp) [leafnode] {$\neg A$};
      
      \node (c1) at (30bp,-75bp)[leafnode] {$C$};
      \node (c0) at (00bp,-75bp) [leafnode] {$\neg C$};
      
      \begin{scope}[on background layer]
        \draw [acarrow] (db0) -- (db);
        \draw [acarrow] (db1) -- (db);
        
        \draw [acarrow] (b0) -- (db0);
        \draw [acarrow] (b1) -- (db1);
        \draw [acarrow] (p8) -- (db1);
        
        \draw [acarrow] (dc) -- (db0);
        \draw [acarrow] (dc0) -- (dc);
        \draw [acarrow] (dc1) -- (dc);
        
        \draw [acarrow] (c0) -- (dc0);
        \draw [acarrow] (c1) -- (dc1);
        \draw [acarrow] (p3) -- (dc0);
        \draw [acarrow] (p8) -- (dc1);
        
        \draw [acarrow] (sc) -- (db1);
        \draw [acarrow] (c0) -- (sc);
        \draw [acarrow] (c1) -- (sc);
        
        \draw [acarrow] (sa) -- (db0);
        \draw [acarrow] (a0) -- (sa);
        \draw [acarrow] (a1) -- (sa);
        
        \draw [acarrow] (da) -- (db1);
        \draw [acarrow] (da0) -- (da);
        \draw [acarrow] (da1) -- (da);
        
        \draw [acarrow] (a0) -- (da0);
        \draw [acarrow] (a1) -- (da1);
        
        \draw [acarrow] (p2) -- (da0);
        \draw [acarrow] (p2) -- (db0);
        \draw [acarrow] (p5) -- (da1);
      \end{scope}
    \end{scope}
    
    \end{tikzpicture}
}
  \caption{Multiplying Arithmetic Circuits: Factor graph and ACs for individual factors which multiply into a single AC for the joint distribution. Given an AC, inference is tractable by propagating inputs.}
  \label{fig:motex1}
\end{figure*}

\section{Collapsed Compilation}
\label{sec:colcomp}
Online collapsed importance sampling presents us with a powerful technique for
adapting to problems traditional collapsed importance sampling may struggle
with. However, it also demands we solve several difficult tasks: one needs a good
proposal distribution over any subset of variables, an efficient way of
computing exactly the probability of a generic query, and an efficient way of
finding the true probability of sampled variables. In this section, we introduce
collapsed compilation, which tackles all three of these problems at once using
techniques from knowledge compilation.

\subsection{Knowledge Compilation Background}

We begin with a short review of how to perform exact inference on a probabilistic
graphical model using knowledge compilation to arithmetic circuits (ACs).
Suppose we have a factor graph consisting of three binary
variables $A$, $B$ and $C$,  and factors $f_1, f_2$ as
depicted in Figure \ref{fig:motex1}. 
Each of these factors, as well as the joint distribution of the product can be represented as an arithmetic circuit. These circuits have inputs corresponding to variable assignments (e.g., $A$ and $\neg A$) or constants (e.g., 5). Internal nodes are sums or products. We can encode a complete assignment by setting the corresponding variable assignments to 1 and the opposing assignments to 0. Then, the root of the circuit evaluates the weight (unnormalized probability) of that world.
Moreover, because of two important properties, we are also able to perform marginal inference.
Product nodes are \emph{decomposable}, meaning that their inputs are disjoint, having no variable inputs in common. Sum nodes are \emph{deterministic}, meaning that for any given complete input assignment to the circuit, at most one of the sum's inputs evaluates to a non-zero value. Now, by setting both assignments for the same variable to 1, we effectively marginalize out that variable. For example, by setting all inputs to 1, the arithmetic circuit evaluates to the partition function of the graphical model.
We refer to \citet{darwiche2009modeling} for further details on how to reason with arithmetic circuits.

In practice, arithmetic circuits are often compiled from graphical models through a logical task called weighted model counting, followed by Boolean circuit compilation techniques. We refer to \citet{Choi2013CompilingPG} and \citet{Chavira2008OnPI} for details.
As our compilation target, we will use the sentential decision diagram (SDD) \citep{Darwiche2011SDDAN}. 
Moreover, given any two SDDs representing factors $f_1, f_2$, and a variable
$\Xr$, we can efficiently compute the SDD
representing the factor multiplication of $f_1$ and $f_2$, as well as the result of conditioning
$f_1$ on $\Xr=x$.  We call such
operations \textbf{apply}, and they are the key to using knowledge compilation
for doing online collapsed importance sampling.
The result of multiplying two arithmetic circuits is depicted in Figure \ref{fig:motex1}.

As a result of SDDs supporting the \textbf{apply} operations, we can directly
compile graphical models to circuits in a bottom-up manner. Concretely, we start out by compiling each factor into a
corresponding SDD representation using the encoding of \citet{Choi2013CompilingPG}. Next, these SDDs are multiplied in order
to obtain a representation for the entire model.
As shown by \citet{Choi2013CompilingPG}, this straightforward
approach can be used to achieve state-of-the-art exact inference on
probabilistic graphical models.

\subsection{Algorithm}
Now that we have proposed online collapsed importance sampling and given
background on knowledge compilation, we are ready to introduce collapsed compilation, an algorithm that uses knowledge compilation to do online collapsed importance sampling. 
Collapsed compilation begins multiplying factors, and when the the SDD becomes too large,
we sample and condition variables until it is sufficiently small again. At the end, the
sampled variables form $\Xp$, and the variables remaining in the SDD form $\Xd$. All SDD \textbf{apply} operations are tractable~\citep{VdBAAAI15b}.

Concretely, collapsed compilation repeatedly performs a few simple steps, following Algorithm~\ref{alg:oncolims}:
\vspace{-0.1cm}
\begin{enumerate}
\itemsep0.10cm 
\item Choose an order, and begin compiling factors until the size limit is reached
\item Select a variable $\Xr_i$ using the given policy $\pi$
\item Sample $\Xr_i$ according to its marginal in the current SDD for the partially compiled graph
\item Condition the SDD on the sampled value for $\Xr_i$
\end{enumerate}

The full technical description and implementation details can be found in Appendix~\ref{a:collcompdet} and~\ref{a:algdetails}.

\section{Experimental Evaluation}
\label{sec:expeval}

\paragraph{Data \& Evaluation Criteria}
To empirically investigate collapsed compilation, we evaluate the performance of
estimating a single marginal on a series of commonly used graphical models. Each model is followed in brackets by the count of nodes and factors.

From the 2014 UAI inference competition, we evaluate on linkage(1077,1077),
Grids(100,300), DBN(40, 440), and Segmentation(228,845) problem instances.
From the 2008 UAI inference competition, we use two semi-deterministic grid
instances, 50-20(400, 400) and 75-26(676, 676). Here the first number indicates the number factor
entries that are deterministic, and the second indicates the size of the
grid.
Finally, we generated a randomized frustrated Ising model on a 16x16 grid,
frust16(256, 480).
Beyond these seven benchmarks, we experimented on ten additional standard benchmarks. Because those were either too easy (showing no difference between collapsed compilation and the baselines), or similar to other benchmarks, we do not report on them here.

For evaluation, we run all sampling-based methods 5 times for 1 hour each. We report the median
Hellinger distance across all runs, which for discrete
distributions $P$ and $Q$ is given by $H(P,Q)=\frac{1}{\sqrt{2}}\sqrt{\sum_{i=1}^k(\sqrt{p_i} - \sqrt{q_i})^2}$.
A * symbol means that the method performed exact inference.

\paragraph{Variable Selection Policies}
We will evaluate the policies detailed in Appendix~\ref{a:algdetails}.
The first policy {\tt RBVar} explores the idea of picking the variable that least increases
the Rao-Blackwell variance of the query \citep{darwiche2009modeling}. For a given
query $\alpha$, to select our next variable from $\Xs$, we use
$\argmin_{\Xr \in \Xs}\sum_{\Xr}P(\alpha | \Xr)^2 P(\Xr)$. This can be computed in time linear
in the size of the current~SDD.

The next policy we look at is {\tt MinEnt}, which selects the variable with
the smallest entropy. Intuitively, this is selecting the variable for which
sampling assumes the least amount of unknown information.

Finally, we examine a graph-based policy {\tt FD} (FrontierDistance). At any given point in our compilation
we have some frontier $\mathcal{F}$, which is the set of variables which have
some but not all factors included in the current SDD. Then we select the
variable in our current SDD which is, on the graph of our model, closest to the
``center'' induced by $\mathcal{F}$. That is, we use $\argmin_{\Xr \in
  \Xs}\max_{\Fr \in \mathcal{F}} ~ \dist(\Xr, \Fr)$.

\subsection{Understanding Collapsed Compilation}

We begin our evaluation with experiments designed to
shed some light on different components involved in collapsed compilation.
First, we evaluate our choice in proposal distribution by comparison to
marginal-based proposals. Then, we examine the effects of setting different
size thresholds for compilation on the overall performance, as well as the sample count and quality.

\paragraph{Evaluating the Proposal Distribution}
Selecting an effective proposal distribution is key to successfully using importance sampling estimation \citep{tokdar2010importance}. As discussed in
Section~\ref{sec:colcomp}, one requirement of online collapsed
importance sampling is that we must provide a proposal distribution
over any subset of variables. Alternatively, this means a proposal for any variable conditioned on any partial instantiation, which is challenging to provide for many traditional choices of proposal.

To evaluate the quality of collapsed compilation's proposal distribution, we
compare it to using marginal-based proposals, and highlight the
problem with such proposals. First, we compare to a dummy uniform proposal. Second, we compare to a proposal
that uses the true marginals for each variable. Experiments on the 50-20 benchmark are shown in Table~\ref{table:proposal}.

\begin{table}[tb]
  \centering
  \caption{Internal comparisons for collapsed compilation. Values represent Hellinger distances.}
  \begin{subtable}{.49\textwidth}
  \caption{Comparison of proposal distributions}
  \label{table:proposal}
  \small
  \scalebox{1.0}{
  \begin{tabular}{l | l | l | l}
    Policy  & Dummy & True & SDD \\
    \hline
    {\tt FD} & \num{2.37e-04} & \num{1.77e-04} & \num{3.72e-07} \\
    {\tt MinEnt} & \num{3.29e-04} & \num{1.31e-03} & \num{2.10e-08} \\
    {\tt RBVar} & \num{5.81e-03} & \num{5.71e-3} & \num{7.34e-03} \\
  \end{tabular}
  }
\end{subtable}%
~
\begin{subtable}{.49\textwidth}
  \centering
  \caption{Comparison of size thresholds}
  \label{table:sizes}
  \small
  \scalebox{1.0}{
  \begin{tabular}{l | l | l | l}
    Policy & 10k & 100k & 1m \\
    \hline
    {\tt FD} & \num{7.33e-5} & \num{3.21e-7} & \num{7.53e-6} \\
    {\tt MinEnt} & \num{1.44e-3} & \num{1.59e-5} & \num{8.07e-4} \\
    {\tt RBVar} & \num{2.96e-2} & \num{2.60e-2} & \num{8.81e-3} \\
  \end{tabular}
  }
\end{subtable}
\\
\begin{subtable}{.48\textwidth}
  \centering
  \caption{Comparison of size thresholds (50 samples)}
  \label{table:fixedsamp}
  \small
  \begin{tabular}{l | l | l | l}
    Policy & 10k & 100k & 1m \\
    \hline
    {\tt FD} & \num{1.63e-3} & \num{5.08e-7} & \num{1.27e-6} \\
    {\tt MinEnt} & \num{1.69e-2} & \num{1.84e-6} & \num{7.24e-6} \\
    {\tt RBVar} & \num{1.94e-2} & \num{1.52e-1} & \num{3.07e-2} \\
  \end{tabular}
\end{subtable}%
~
\begin{subtable}{.48\textwidth}
  \centering
  \caption{Number of samples taken in 1 hour by size}
  \label{table:numsamp}
  \small
  \begin{tabular}{l | l | l | l}
    Size Threshold & 10k & 100k & 1m \\
    \hline
    Number of Samples & 561.3 & 33.5 & 4.7 \\
  \end{tabular}
  \vspace{0.69cm}
\end{subtable}
\end{table}

Particularly with policies {\tt FrontierDist} and {\tt
  MinEnt}, the results underline the effectiveness of collapsed compilation's proposal distribution over baselines. This is the effect of conditioning --
even sampling from the true posterior marginals does not work very well, due to
the missed correlation between variables. Since we are already conditioning for our partial exact inference, collapsed compilation's proposal distribution is providing this improvement for very little added cost.

\paragraph{Choosing a Size Threshold}
A second requirement for collapsed compilation is to
set a size threshold for the circuit being maintained. Setting the threshold
to be infinity leaves us with exact inference which is in general intractable,
while setting the threshold to zero leaves us with importance sampling using what is
likely a poor proposal distribution (since we can only consider one factor at a
time). Clearly, the optimal choice finds a trade-off between these two considerations.

Using benchmark 50-20 again, we compare the performance on three different settings
for the circuit size threshold: 10,000, 100,000, and 1,000,000.
Table~\ref{table:sizes} shows that generally, 100k gives the best performance,
but the results are often similar. To further investigate this, Table~\ref{table:fixedsamp} and
Table~\ref{table:numsamp} show performance with exactly 50 samples for each
size, and number of samples per hour respectively. This is more informative as
to why 100k gave the best performance - there is a massive difference in
performance for a fixed number of samples between 10k and 100k or 1m. The gap
between 100k and 1m is quite small, so as a result the increased samples for
100k leads to better performance. Intuitively, this is due to the nature of exact circuit compilation, where at a certain size point of compilation you enter an exponential regime. Ideally, we would like to stop compiling right before we reach that point. Thus, we proceed with 100k as our size-threshold setting for further experiments.



\subsection{Memory-Constrained Comparison}
In this section, we compare collapsed compilation to two related state-of-the-art methods:
edge-deletion belief propagation (EDBP) \citep{choi2006edge}, and IJGP-Samplesearch (SS)
\citep{gogate2011samplesearch}. Generally, for example in past UAI probabilistic inference competitions, comparing methods in this space
involves a fixed amount of time and memory being given to each tool.
The results are then directly compared to determine the empirically best
performing algorithm. While this is certainly a useful metric, it is highly
dependent on efficiency of implementation, and moreover does not provide as good
of an understanding of the effects of being allowed to do more or less exact inference. To give more informative results, in addition to a time limit, we restrict our comparison at the algorithmic level, by controlling for the level of exact inference being performed.

\paragraph{Edge-Deletion Belief Propagation}
EDBP performs approximate inference by increasingly running more exact junction tree inference, and approximating the rest via belief propagation \citep{choi2006edge,choi2005edbp}. To constrain EDBP, we limit the corresponding circuit size for the junction tree used. In our experiments we set these limits at 100,000 and 1,000,000.

\paragraph{IJGP-Samplesearch}
IJGP-Samplesearch is an importance sampler
augmented with constraint satisfaction search \citep{gogate2011samplesearch,gogate2007samplesearch}. It uses iterative join graph propagation
\citep{dechter2002iterative} together with $w$-cutset sampling
\citep{bidyuk2007cutset} to form a proposal, and then uses search
to ensure that no samples are rejected. To constrain SS, we limit treewidth $w$ at either 15, 12, or 10. For reference, a circuit of size 100,000 corresponds to a treewidth between 10 and 12.

Appendix~\ref{a:exp} describes both baselines as well as the experimental setup in further detail.

\begin{table}[tb]
  \small
  \caption{Hellinger distances across methods with internal treewidth and size bounds \label{table:limitedsize}}
  \vspace{0.08cm}
  \centering
  \begin{tabular}{l|l | l |l |l|l|l|l}
    Method & 50-20 & 75-26 & DBN & Grids & Segment & linkage & frust \\
    \hline
    {\tt EDBP-100k} & \num{2.19e-3} & \num{3.17e-5} & \num{6.39e-1} & \num{1.24e-3} & \num{1.63e-6} & \num{6.54e-8} & \num{4.73e-3} \\
    {\tt EDBP-1m} & \num{7.40e-7} & \num{2.21e-4} & \num{6.39e-1} & \num{1.98e-7} & \num{1.93e-7} & \num{5.98e-8} & \num{4.73e-3} \\
    \hline
    \hline
    {\tt SS-10} & \num{2.51e-2} & \num{2.22e-3} & \num{6.37e-1} & \num{3.10e-1} & \num{3.11e-7} & \num{4.93e-2} & \num{1.05e-2} \\
    {\tt SS-12} & \num{6.96e-3} & \num{1.02e-3} & \num{6.27e-1} & \num{2.48e-1} & \num{3.11e-7} & \num{1.10e-3} & \num{5.27e-4}\\
    {\tt SS-15} & \num{9.09e-6} & \num{1.09e-4} & * & \num{8.74e-4} & \num{3.11e-7} & \num{4.06e-6} & \num{6.23e-3} \\
    \hline
    \hline
    {\tt FD} & \num{9.77e-6} & \num{1.87e-3} & \num{1.24e-1} & \num{1.98e-4} & \num{6.00e-8} & \num{5.99e-6} & \num{5.96e-6} \\
    {\tt MinEnt} & \num{1.50e-5} & \num{3.29e-2} & \num{1.83e-2} & \num{3.61e-3} & \num{3.40e-7} & \num{6.16e-5} & \num{3.10e-2} \\
    {\tt RBVar} & \num{2.66e-2} & \num{4.39e-1} & \num{6.27e-3} & \num{1.20e-1} & \num{3.01e-7} & \num{2.02e-2} & \num{2.30e-3} \\
          \end{tabular}
\end{table}

\subsubsection{Discussion}

Table~\ref{table:limitedsize} shows the experimental results for this setting.
Overall, we have found that when restricting all methods to only do a fixed
amount of exact inference, collapsed compilation has similar performance to both
samplesearch and EDBP. Furthermore, given a good choice of variable selection policy, it can
often perform better. In particular, we highlight DBN, where we see that
collapsed compilation with the {\tt RBVar} or {\tt MinEnt} policies is the only method that
manages to achieve reasonable approximate inference. This follows the intuition discussed in Section~\ref{s:motivation}: a good choice of a few variables in a densely connected model can lead to relatively easy exact inference for a large chunk of the model.

Another factor differentiating collapsed compilation from both EDBP and
samplesearch is the lack of reliance on some type of belief propagation
algorithm. Loopy belief propagation is a cornerstone of approximate inference in
graphical models, but it is known to have problems converging to a good
approximation on certain classes of models \citep{murphy1999loopy}. The problem
instance frust16 is one such example - it is an Ising model with spins set up
such that potentials can form loops, and the performance of both EDBP and
samplesearch highlight these issues.

\subsection{Probabilistic Program Inference}
As an additional point of comparison, we introduce a new type of benchmark. We use the probabilistic logic programming language ProbLog \citep{de2015probabilistic} to model a graph with probabilistic edges, and then query for the probability of two nodes being connected. This problem presents a unique challenge, as every non-unary factor is deterministic.

\begin{wraptable}[9]{R}{.25\textwidth}
\vspace{-0.5cm}
\begin{flushright}
\begin{minipage}{.98\linewidth}
\small
    \begin{tabular}{l|l}
      Method&Prob12 \\
      \hline
 {\tt EDBP-1m} & \num{3.18e-01} \\
 \hline
 \hline
 {\tt SS-15} & \num{3.87e-03} \\     
 \hline
 \hline
 {\tt FD} & \num{1.50e-03} \\
    \end{tabular}
    \caption{Hellinger distances for ProbLog}
  \label{tb:problog}
        \end{minipage}
  \end{flushright}
\end{wraptable}

Table~\ref{tb:problog} shows the results for this benchmark, with the underlying graph being a 12x12 grid. We see that EDBP struggles here due to the large number of deterministic factors, which stop belief propagation from converging in the allowed number of iterations. Samplesearch and collapsed compilation show similarly decent results, but interestingly they are not happening for the same reason. To contextualize this discussion, consider the stability of each method. Collapsed compilation draws far fewer samples than SS - some of this is made up for by how powerful collapsing is as a variance reduction technique, but it is indeed less stable than SS. For this particular instance, we found that while different runs for collapsed compilation tended to give different marginals fairly near the true value, SS consistently gave the same incorrect marginal. This suggests that if we ran each algorithm until convergence, collapsed compilation would tend toward the correct solution, while SS would not, and appears to have a bias on this benchmark.
  

\section{Related Work and Conclusions}

We have presented online collapsed importance sampling, an asymptotically
unbiased estimator which allows for doing collapsed importance sampling without
choosing which variables to collapse a priori. Using techniques from knowledge
compilation, we developed collapsed compilation, an implementation of online
collapsed importance sampling that draws its proposal distribution from partial
compilations of the distribution, and naturally exploits structure in the distribution.

In related work, \citet{lowd2010approximate} study arithmetic circuits as a variational approximation of graphical models. Approximate compilation has been used for inference in probabilistic (logic) programs \citep{494681}. Other approximate inference algorithms that exploit local structure include samplesearch and the family of universal hashing algorithms \citep{ermon2013embed,chakraborty2014distribution}.
Finally, collapsed compilation can be viewed as an
approximate knowledge compilation method: each drawn sample presents a
partial knowledge base along with the corresponding correction weight. This means
that it can be used to approximate any query which can be performed efficiently on an
SDD -- for example the most probable explanation (MPE) query \citep{heichan2006,choi2017relaxing}. We leave this as an interesting direction for future work.


\clearpage


%

\bibliography{reference}
\bibliographystyle{icml2018}

\clearpage

\appendix
\section{Proof of Theorems}\label{a:proofs}
We begin with the formal definition of our augmented factor graph.
\begin{definition}
  Suppose we have a discrete distribution represented by a factor graph $F$,
  with variables $\Xs$. Then we define the corresponding augmented factor graph
  $F_A$ as follows:
  \begin{itemize}
  \item For each $\Xr_i$, we introduce variables $\Xr^*_i$ and $S_i$
  \item For each $\Sr_i$, add a unary factor such that $\Sr_i$ is uniform
  \item For each $\Xr^*_i$, add a factor such that
    \begin{align}
   P(\Xr_i^*|\Sr_i,\Xr_i) =  \begin{cases}
    0.5, & \text{if } \Sr_i=0 \\
    1, & \text{if } \Sr_i=1,\Xr_i=\Xr_i^* \\
    0, & \text{otherwise}
  \end{cases}
         \end{align}

  \end{itemize}
\end{definition}
\subsection{Proof of Theorem~\ref{thm:equivaug}}
\begin{proof}
  Consider some variable $\Xr_i$ together with its corresponding augmented
  variables $\Xr^*_i, \Sr_i$. Then examining the resulting graphical model, we
  see that $\Xr*_i$ and $\Sr_i$ are only connected to the rest of the model via
  $\Xr_i$. Due to the conditional independence properties, this means that we
  can first sum out $\Sr_i$ from $P(\Xr^*_i|\Sr_i, \Xr_i)$, and then simply sum
  $\Xr^*_i$. Thus, the result of any query over $\Xs$ on $F_A$ will be
  equivalent to the result on $F$, as we can simply sum out all additional
  variables, and end up at the same model.
\end{proof}
\subsection{Proof of Theorem~\ref{thm:onlest}}
\begin{proof}
  Our goal is to estimate
\begin{align*}
  \label{eq:5}
  \mathbb{E}_{P(\xi)}[f(\xi)]&=\sum_{X, X^*, S}P(X,X^*,S)f(X) \\
  &=\sum_{X, X^*, S}P(X)\prod_iP(S_i)P(X^*_i|X_i,S_i)
\end{align*}
We begin by introducing our proposal distribution $Q(X^*,S)$, and supposing we have samples $\{X^*[m], S[m]\}_{m=1}^M$ drawn from our proposal
distribution $Q$. Now, substituting this in and continuing our previous
equations gives us:
\begin{align*}
  &\sum_{X^*[m],S[m]}\frac{1}{Q(X^*[m],S[m])} \\
  &(\sum_XP(X)\prod_iP(S_i[m])P(X_i^*[m])f(X))
\end{align*}

We can split our sum over $X$ up into indices where $S_i[m] = 0$, and ones where
$S_i[m]=1$. For the sake of notation, we will refer to these sets of indices as
$I^C$ and $I$ respectively:
\begin{align*}
  &\sum_{m=1}^M\frac{1}{Q(X^*[m],S[m])}\\&\sum_{X_I}\prod_{i \in I}(P(S_i[m])P(X_i^*[m]|X_i,S_i[m]))\\  & \sum_{X_{I^C}}P(X)\prod_{i \in I^C}P(S_i[m])P(X_i^*[m]|X_i,S_i[m])f(X) \\
  =& \sum_{m=1}^M\frac{1}{Q(X^*[m],S[m])} \\ &\sum_{X_I}\prod_{i \in I}(P(S_i[m])P(X_i^*[m]|X_i,S_i[m]))\\ &\sum_{X_{I^C}}P(X_I)P(X_{I^C}|X_I) \\ & \prod_{i \in I^C}(0.5)P(X_i^*[m]|X_i,S_i[m])f(X) \\
  =& \sum_{m=1}^M\frac{1}{Q(X^*[m],S[m])} \\ & \sum_{X_I}P(X_I)\prod_{i^I}(P(S_i[m])P(X_i^*[m]|X_i,S_i[m]))\\ & \sum_{X_{I^C}}P(X_{I^C}|X_I)(0.5*0.5)^{|I^C|}f(X) \\
  =&\sum_{m=1}^M\frac{(0.25)^{|I^C|}}{Q(X^*[m],S[m])}\sum_{X_I}P(X_I)\\ & \prod_{i \in I}(0.5)P(X_i^*[m]|X_i,S_i[m])\sum_{X_{I^C}}P(X_{I^C}|X_I)f(X) \\
  =& \sum_{m=1}^M\frac{(0.25)^{|I^C|}}{Q(X^*[m],S[m])}(0.5)^{|I|}\sum_{X_I}P(X_I)\\ & \prod_{i \in I}P(X_i^*[m]|X_i,S_i[m])\mathbb{E}_{P(X_{I^C}|X_I}[f(X)] \\
\end{align*}

Now, observe that the term $\prod_{i \in I}P(X_i^*[m]|X_i,S_i[m]) = 1 \iff
\forall i \text{ s.t. } S_i[m]=1, X_i=X^*_i[m]$. Since there is only one setting
of these indices that satisfies this (that is, letting $X_i=X^*_i[m]$ everywhere
that $S_i[m]=1$), this allows us to get rid of this sum, and our equation
becomes:
\begin{align*}
  \sum_{m=1}^M\frac{(0.25)^{I^C}(0.5)^{I}P_{X_I}(X_I^*[m])\mathbb{E}_{P(X_{I^C}|X_I)}[f(X)]}{Q(X^*[m],S[m])} \\
\end{align*}

Which is precisely what we wanted - this is the equation we would expect to use
for our online collapsed importance sampler (once we adjust our proposal
distribution for variables that are not actually sampled to correct for the
$0.5$s that are left).
\end{proof}

\subsection{Proof of Theorem~\ref{thm:prop}}
\begin{proof}
  We will proceed by contradiction. Suppose we have two paths through our
  variables, $X^*_{i_1},X^*_{i_2},\dots,X^*_{i_n}$ and
  $X^*_{j_1},X^*_{j_2},\dots,X^*_{j_n}$ which can produce the same assignment to
  all variables. Now, there are two key facts we will observe and make use of:

  \begin{enumerate}
  \item The starting point for any path is fixed, that is $X^*_{i_1}=X^*_{j_1}$.
    Our heuristics are deterministic, and the memory limit remains constant, so
    the first variable to sampled is always the same.
  \item Once an assignment is made to the current variable being sampled, the
    decision of which variable to sample is deterministic - again, because our
    heuristics must be deterministic. To put it another way, if
    $x^*_{i_k}=x^*_{j_k}$, then $X^*_{i_k+1}=X^*_{j_k+1}$.
  \end{enumerate}
  Now, since path $i$ and path $j$ are different, there must be some first
  element $k$ where $X^*_{i_k}\ne X^*_{j_k}$. By fact 1, $k>1$. Also, observe
  that since $k$ is the first such element, $X^*_{i_{k-1}}=X^*_{j_{k-1}}$. But
  since our 2 paths must give the same assignment to all variables, this means
  also that $x^*_{i_{k-1}}=x^*_{j_{k-1}}$, which means that
  $X^*_{i_k}=X^*_{j_k}$ by fact 2. This is a contradiction.

\end{proof}

\section{Collapsed Compilation: Algorithm Outline}\label{a:collcompdet}
Algorithm~\ref{alg:colcomp} describes Collapsed Compilation in more detail. Note that to
avoid confusion, we hereafter refer to the processing of sampling $x_i \sim
Q(\Xr_i)$ as conditioning (since all future variables sampled are conditioned on
$x_i$), and a single full run as a sample.

\begin{algorithm2e}
  \caption{A single sample of Collapsed Compilation}
  \label{alg:colcomp}
  \SetKwInOut{Input}{Input}
  \DontPrintSemicolon
  \Input{$\pi$: A variable selection policy computed using an SDD,\\
    $M$: A probabilistic graphical model, \\
  $f$: Function to estimate expectation for}
  \KwResult{A single sample $\Xpm[m], \mathbb{E}_{P(\Xdm | \xpm[m])}[f(\xpm[m],
    \Xdm)], w[m]$}
  $\mathit{SDD} \leftarrow \mathit{True}$ \;
  $\Xp \leftarrow \{\}$ \;
  $q \leftarrow 1$ \;
  \While{$M$ is not compiled}{
    $\mathit{SDD} \leftarrow \mathit{bottomUpCompileStep(\mathit{SDD}, M)}$ \;
    \While{SDD is too large} {
      $\Xr_j \leftarrow \pi(\mathit{SDD}, \Xp)$ 
      $x_j  \sim P_{\mathit{SDD}}(\Xr_j)$ \;
      $q \leftarrow q * P_{\mathit{SDD}}(x_j)$ \;
      $\Xp \leftarrow \Xp \bigcup {\Xr_j=x_j}$ \;
      $\mathit{SDD} \leftarrow \mathit{SDD}|x_j$ \;
    }
  }
  \Return{$\Xp, \frac{\mathit{WMC}_f(\mathit{SDD})}{\mathit{WMC}(\mathit{SDD})}, \frac{q}{\mathit{WMC}(\mathit{SDD})}$}
\end{algorithm2e}

There are a few important things to take away here. First, if we at any point
interrupt bottom-up compilation, what we will get is a complete compilation of
some subset of the model. This means that on line 7, the proposal distribution
$P_{SDD}$ we are drawing from is the true distribution for $X_j$ on some subset of $M$,
conditioned on all previous variables.

Secondly, there are a few calls to a weighted model count function $\mathit{WMC}$ on line
11. Recall that for an SDD representing a probability distribution
$P(\Xs)$, the weighted model count subject to a function $f$ computes
$\sum_{\xs}P(\xs)f(\xs)$. Also, observe that the SDD we are left with when we
finish compiling is representing the joint distribution $P(\Xd, \Xp=\xp)$. Thus,
observing that $P(\Xd, \Xp=\xp) = \hat{P}(\Xd | \Xp=\xp)$
we see that the weighted model count subject to $f$ is actually
$\sum_{\xd}\hat{P}(\Xd=\xd| \Xp=\xp)f(\xd, \xp)$. But setting $f\equiv 1$ allows
us to compute the normalization constant, meaning that
$\frac{\mathit{WMC}_f(\mathit{SDD})}{\mathit{WMC}(\mathit{SDD})}=\mathbb{E}_{P(\Xd | \Xp=\xp)}[f(\xp, \Xd)]$.

\section{Collapsed Compilation: Algorithmic Details}\label{a:algdetails}
There are many moving parts in this method, so in this section we will examine and
each in isolation.
\subsection{Compilation Order}
Once we have compiled an SDD for each factor in the graph, bottom up compilation
allows us to choose in which order to multiply these SDDs. In our experiments, we
look at two orders: {\tt BFS} and {\tt revBFS}. The first begins from the
marginal query variable, and compiles outwards in a breadth first order. The
second does the same, but in exactly the opposite order arriving at the query
variable last.
\subsection{Proposal Distribution}
Given that we have decided to condition on a variable $\Xr_j$, we decide its
proposal distribution by computing the marginal probability of $\Xr_j$ in our
currently compiled SDD. This can be done in time linear in the size of the
circuit by computing the partial derivative of $\Xr_j$ with respect to the
weighted model count in the current circuit \citep{Darwichediff2003}.
\subsection{Variable Selection}
The manner in which we select the next variable to condition on - that is, our
choice of {\tt nextVar} in algorithm \ref{alg:colcomp} - has a large effect on
both the tractability and efficiency of our sampler. 
We will explore three policies, the first of which depends specifically on the
marginal being queried for, while the other two do not. All of the policies we
explore will satisfy Definition~\ref{def:det} - that is, they are all deterministic.  

The first policy {\tt RBVar} explores the idea of picking the variable which least increases
the Rao-Blackwell variance of the query \citep{darwiche2009modeling}. For a given
query $\alpha$ to select our next variable from $\Xs$ we use
$\argmin_{\Xr \in \Xs}\sum_{\Xr}P(\alpha | \Xr)^2 P(\Xr)$. This can be computed in time linear
in the size of the current SDD.

The next policy we look at is {\tt MinEnt}, which selects the variable with
the smallest entropy. Intuitively, this is selecting the variable for which
sampling it is assuming the least unknown information.

Finally, we examine a graph based policy {\tt FrontierDist}. At any given point in our compilation
we have some frontier $\mathcal{F}$, which is the set of variables which have
some but not all factors included in the current SDD. Then we select the
variable in our current SDD which is, on the graph of our model, closest to the
``center'' induced by $\mathcal{F}$. That is, we use $argmin_{\Xr \in
  \Xs}\max_{\Fr \in \mathcal{F}}dist(\Xr, \Fr)$.

In our experiments, policy {\tt RBVar} is used with the compilation order {\tt
  BFS}, while policies {\tt MinEnt} and {\tt FrontierDist} are used with {\tt RevBFS}.

\subsection{Determinism}
A desirable property for samplers - particularly when there are a large number of
deterministic relationships present in the model - is to be
\emph{rejection-free}. It is clear that in the presence of
no deterministic factors (that is, no 0 entries in any factor), collapsed
compilation will never reject samples. Here, we describe how this guarantee can be
maintained in the presence of 0-valued factor entries.
\paragraph{Extracting a Logical Base}
\label{sec:extractbase}
Suppose we are given a factor over some variables $\Xr_1,\Xr_2...,\Xr_n$. Then a
weight of 0 given to an assignment $x_1,x_2..,x_n$ indicates a logical statement.
Specifically, we can say for certain over the entire model that $\neg (x_1\land
x_2,...,\land x_n)$. As a preprocessing step, we find all such factor entries in
the graph, convert them each into the corresponding logical statement, and then
take the conjunction of all of these. This forms the logical base for our model.

\paragraph{An Oracle for Determinism}
Once we have obtained this base for our model, we can precompile a circuit
representing it. This allows us to make queries asking whether
there is a non-zero probability that $\Xr_j=x_j$, given all previous assignments
$\Xp$~\citep{Darwiche1999CompilingKI}. Thus, when we sample from the marginal
probability of $\Xr_j$ from our current SDD (our proposal distribution),
we first renormalize this marginal to only include assignments which have a
non-zero probability according to our determinism oracle. Of course, this is not
always possible due to size constraints in the case where there is an enormous
amount of determinism. For these cases we just run collapsed compilation as is -
depending on the compilation order it will still tend to reject few samples.
\paragraph{Literal Entailment}
As a simple optimization, we can recognize any
variables whose values are already deterministically chosen based on previously
conditioned variables, and assign them as such in our SDD. Given a determinism
oracle, deciding this can be done for all variables in the model in time linear
in the size of the oracle \citep{Darwiche2001DecomposableNN}.

\section{Experimental Details} \label{a:exp}

\subsection{Edge-Deletion Belief Propagation}
Edge-deletion belief propagation (EDBP) is a method for doing approximate
graphical model inference by using a combination of exact inference and belief
propagation \citep{choi2006edge} \citep{choi2005edbp}. EDBP iteratively computes
more and more of the model exactly using junction tree, at each step performing
belief propagation to approximate the rest of the model. It can be viewed as the
belief propagation analog of collapsed compilation, which makes it an
interesting target for comparison. A major conceptual difference between the two
is that while collapsed compilation is asymptotically unbiased and thus will
given an accurate result given enough time, EDBP will tend to finish more
quickly but has no way to improve once converged.

To capture a more direct comparison of the amount of exact inference being
performed, we compare collapsed compilation to EDBP with the size of the
junction tree used directly being limited, rather than the computer memory
usage. In particular, we limit the size of the circuit corresponding to the
junction tree to be similar to the sizes used for collapsed compilation. To this
end, we use 100,000 and 1,000,000 as size limits for the junction tree, and
otherwise run the algorithm as usual. Table~\ref{table:limitedsize} shows the
results across all benchmarks. Keeping in mind that all policies for collapsed
compilation use 100,00 as their size limit, collapsed compilation is comparable
to EDBP. Both perform very well in linkage and Segment, and while collapsed
compilation performs better on 50-20, EDBP does better on 75-26. 

 \subsection{SampleSearch}
IJGP-Samplesearch is an importance sampler
augmented with constraint satisfaction search \citep{gogate2011samplesearch}
\citep{gogate2007samplesearch}. It uses iterative join graph propagation
\citep{dechter2002iterative} together with $w$-cutset sampling
\citep{bidyuk2007cutset} to form a proposal, and then uses search
to ensure that no samples are rejected.

Once again, we would like to control for the amount of exact inference being
done directly at the algorithmic level, rather than via computer memory. For samplesearch,
we do this by limiting $w$, which is the largest treewidth that is allowed when
using collapsing to reduce variance. We run samplesearch with three different
settings, limiting $w$ to 15, 12, and 10 respectively.
Table~\ref{table:limitedsize} shows the results of running our standard set of
benchmarks with all of these settings. As a reference point, empirically a circuit
size limit of 100,00 generally corresponds to a treewidth somewhere between 10
and 12. The results are generally similar to constraining the memory of EDBP,
but with more constrained versions of samplesearch suffering more. For example,
although linkage appears to be an easy instance in general, without a large
enough $w$-cutset, samplesearch struggles compared to other methods.

\end{document}